\documentclass[sigconf,screen,natbib=true,anonymous=false]{acmart}
\AtBeginDocument{%
  }

\usepackage{afterpage}
\usepackage{multirow}
\usepackage{makecell}
\usepackage{tabularx}
\setcopyright{acmlicensed}
\copyrightyear{2018}
\acmYear{2018}
\acmDOI{XXXXXXX.XXXXXXX}
\acmConference[48th International ACM SIGIR Conference on Research and Development in Information Retrieval]{Make sure to enter the correct
  conference title from your rights confirmation email}{July 13-18, 2025}{Padua, Italy}
\acmISBN{978-1-4503-XXXX-X/2018/06}

\begin{document}

\title{Modelship Attribution: Tracing Multi-Stage Manipulations \\Across Generative Models}

\author{Zhiya Tan}
\affiliation{%
  \institution{College of Computing and Data Science, \\
  Nanyang Technological University}
  \country{Singapore}
}

\author{Xin Zhang}
\affiliation{%
  \institution{Centre for Frontier AI Research,\\
  Agency for Science, Technology and Research}
  \country{Singapore}
}

\author{Joey Tianyi Zhou}
\affiliation{%
  \institution{Centre for Frontier AI Research,\\
  Agency for Science, Technology and Research}
  \country{Singapore}
}
\renewcommand{\shortauthors}{Trovato et al.}

\begin{abstract}
As generative techniques become increasingly accessible, authentic visuals are frequently subjected to iterative alterations by various individuals employing a variety of tools. Currently, to avoid misinformation and ensure accountability, a lot of research on detection and attribution is emerging. Although these methods demonstrate promise in single-stage manipulation scenarios, they fall short when addressing complex real-world iterative manipulation. In this paper, we are the first, to the best of our knowledge, to systematically model this real-world challenge and introduce a novel method to solve it. We define a task called "\textit{\textbf{Modelship Attribution}}", which aims to trace the evolution of manipulated images by identifying the generative models involved and reconstructing the sequence of edits they performed. 
To realistically simulate this scenario, we utilize three generative models, StyleMapGAN, DiffSwap, and FacePartsSwap, that sequentially modify distinct regions of the same image. This process leads to the creation of the first modelship dataset, comprising 83,700 images (16,740 images × 5).
Given that later edits often overwrite the fingerprints of earlier models, the focus shifts from extracting blended fingerprints to characterizing each model's distinctive editing patterns. To tackle this challenge, we introduce the modelship attribution transformer (MAT), a purpose-built framework designed to effectively recognize and attribute the contributions of various models within complex, multi-stage manipulation workflows.
Through extensive experiments and comparative analysis with other related methods, our results, including comprehensive ablation studies, demonstrate that the proposed approach is a highly effective solution for modelship attribution.
\end{abstract}



\begin{teaserfigure}
  \includegraphics[width=\textwidth]{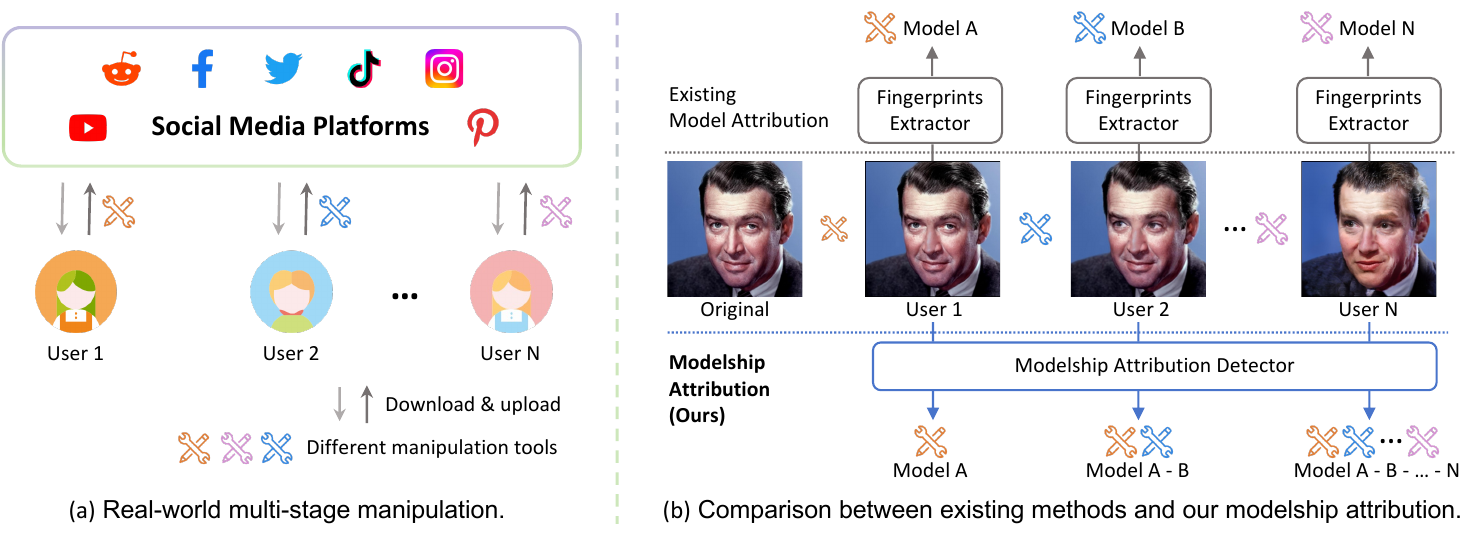}
  \caption{(a) The data shared on social media platforms often undergoes multi-stage manipulation by various individuals utilizing different tools, further complicating the detection and attribution. (b) To address this real-world challenge, unlike previous model attribution methods that focus on extracting specific fingerprints of individual models, we propose modelship attribution, which aims to trace the evolution of the manipulation process.}
  \label{fig:Overview}
\end{teaserfigure}


\maketitle

\section{Introduction}
The advancement of artificial intelligence-generated content (AIGC) technology has elevated generative models to produce highly realistic content \cite{peng2024synctalk, caliskan2025pav}, indistinguishable from authentic media, posing significant risks to information security, public trust, and privacy \cite{guo2024domain, zou2023universal}. The pervasive use of social media amplifies this challenge, as maliciously generated content can be easily uploaded, modified, and repeatedly redistributed by users (shown in \autoref{fig:Overview}(a)). During this process, the content frequently undergoes multi-stage manipulation using diverse generative tools for face replacement, style alterations, and other modifications. This intricate evolution of manipulation makes malicious content increasingly deceptive and difficult to detect, further amplifying cybersecurity risks.

Model attribution is a vital tool for mitigating the spread of synthetic content by identifying generative models based on embedded features and traces. However, existing methods, designed for simplified settings, often fail to generalize effectively in complex, real-world scenarios. On the one hand, these methods are poorly adapted to different types of generative models. 
For example, methods such as AttNet \cite{yu2019attributing} and DNA-Det \cite{yang2022deepfake} are mainly designed for Generative adversarial networks (GANs) \cite{goodfellow2020generative}, and the detection performance drops significantly when confronted with Variational Autoencoders (VAEs) \cite{kingma2013auto} or current state-of-the-art Diffusion Models (DMs) \cite{ho2020denoising}. This limitation of model adaptation makes it difficult for attribution methods to cope with rapidly evolving and diverse generation models. Although the reverse-engineering method proposed by Wang et al. applies to various models—including DCGAN \cite{radford2015unsupervised}, VAE \cite{kingma2013auto}, StyleGAN2-ADA \cite{Karras2020ada}, Consistency Model \cite{song2023consistency}, and ControlGAN \cite{li2019controllable}—its reliance on pretrained weights and architectural details within a white-box reconstruction framework limits its applicability, particularly for black-box generative models.
On the other hand, existing methods also face challenges with multi-stage manipulations, where malicious content undergoes repeated processing, gradually eroding the original model’s fingerprints. These approaches often frame the attribution task as multi-class or binary classification, aiming to identify a single model (as illustrated in \autoref{fig:Overview}(b)). Consequently, their performance deteriorates significantly when handling multiple manipulations, failing to capture the entire sequence of models and edits involved. This limitation in generalizing to diverse generative models and complex manipulations undermines the reliability of existing model attribution techniques in real-world scenarios.

\begin{figure}[t]
  \centering
  \includegraphics[width=\linewidth]{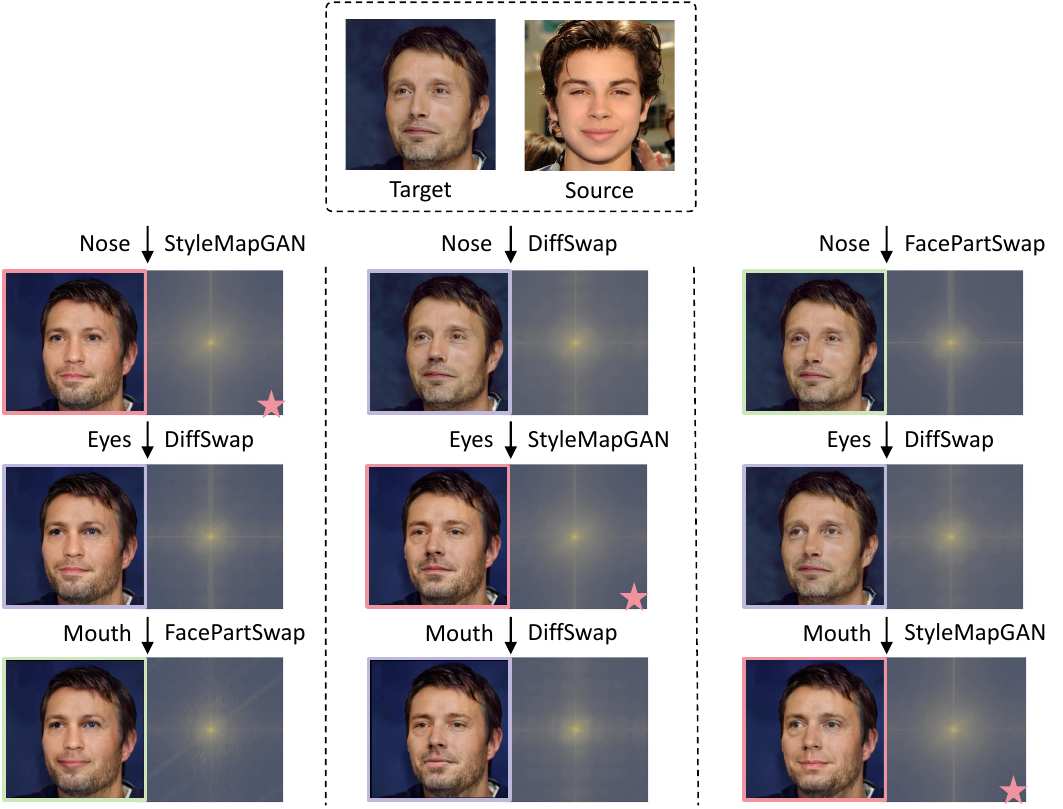}
  \caption{The results show that different models leave distinct "fingerprints": DiffSwap (purple border) \cite{zhao2023diffswap}, FacePartSwap (green border) \cite{ferrari2022makes}, and StyleMapGAN (red border) \cite{kim2021exploiting}. StyleMapGAN's spectra  are highly similar (marked with red stars), essentially covering the fingerprints of earlier stages. So it is more effective to study the editing paradigms of the models. }
  \Description{Examples}
  \label{fig:contrast}
\end{figure}

\autoref{fig:contrast} shows the stage-by-stage results of swapping the nose, eyes, and mouth from the source image onto the target image using different models. The spectra of images generated by different neural networks vary significantly \cite{frank2020leveragingfrequencyanalysisdeep}. To analyze this, we extracted the frequency-domain spectra of each image, which are displayed on the right side of the corresponding results in the figure.
By observing these spectra, it becomes clear that different models leave distinct "fingerprints": among the 9 images, the 4 generated by DiffSwap (purple border) exhibit spectra with numerous bright spots and a wide distribution of energy, indicating that these images are more complex and contain more high-frequency details. The 2 images generated by FacePartSwap (green border) display discrete, multi-directional spectral components. Meanwhile, the 3 images generated by StyleMapGAN (red border) have spectra with higher brightness concentrated in the center, suggesting a smoother overall result.

Further comparison of the 3 StyleMapGAN results across different stages reveals that their spectra are highly similar (marked with red stars), essentially covering the fingerprints of earlier stages. This implies that relying solely on extracted fingerprints to attribute images to specific models is not feasible, as subsequent operations tend to overlay and obscure the traces left by earlier models.
Although the same editing regions were specified at each stage, the final images generated by different models differ significantly. This suggests that a more effective approach would involve tracking the operational traces left on the images to learn the unique editing paradigms of each model.

\begin{figure*}[t]
    \centering
    \includegraphics[width=0.8\textwidth]{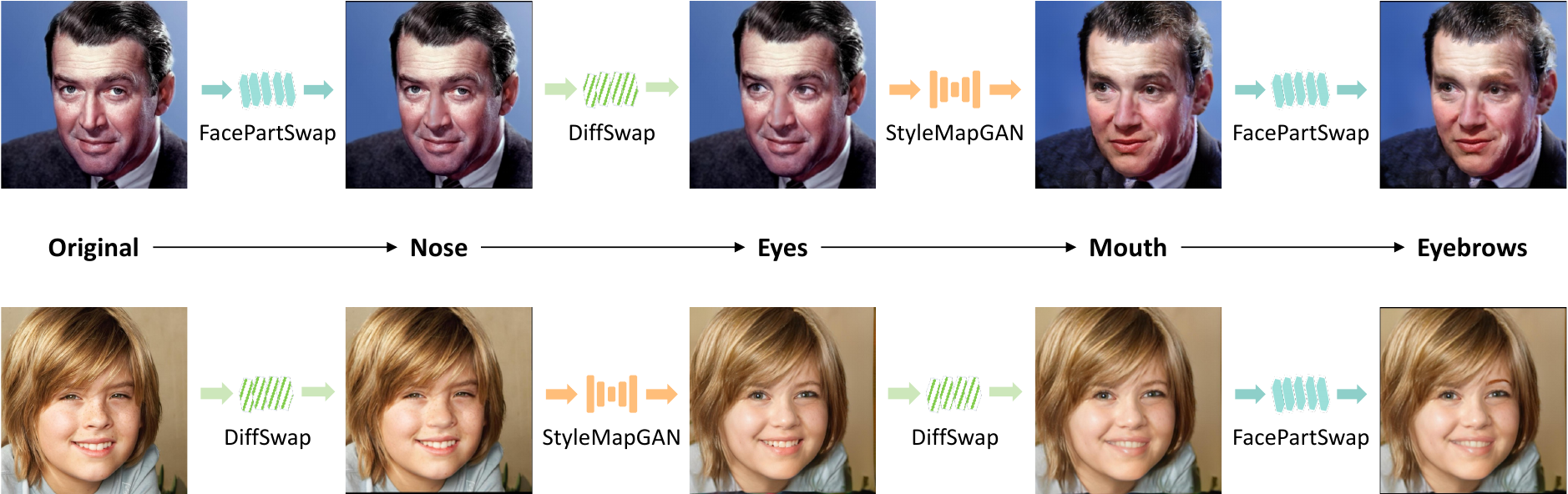}
    \caption{Overview of the modelship dataset generation process. Facial components (nose, eyes, mouth, eyebrows) are sequentially manipulated using three face part-swapping methods: StyleMapGAN \cite{kim2021exploiting}, DiffSwap \cite{zhao2023diffswap}, and FacePartsSwap \cite{ferrari2022makes}. Manipulation order is randomized, and all models are applied across four stages to ensure diverse combinations.}
     \label{fig:dataset}
\end{figure*}

In order to bring the Model Attribution technique closer to complex real-world scenarios, this paper proposes a completely new task, Modelship Attribution.
When an image has been edited in multiple rounds by multiple deepfake tools, Modelship Attribution is not only able to identify the generative model initially used for the image, but also reconstruct all the editing sequences that have been performed, thus tracking the complete evolution of the image. "Modelship" is a newly coined term that denotes the “identity attribution” of one or more models in the process of content creation and editing, as well as the evolution of that process. The reason for calling it “Modelship” is to underscore the model’s role as an “author/owner/contributor” in content generation, thereby facilitating subsequent identification, attribution, and provenance research.

To simulate a more realistic application environment, this paper constructs, for the first time, a Modelship deepfake dataset that selects three deepfake face swapping methods with completely different architectures: StyleMapGAN \cite{kim2021exploiting} (based on GAN), DiffSwap \cite{zhao2023diffswap} (based on DM), and FacePartsSwap \cite{ferrari2022makes} (based on 3D reconstruction). For the same target image, the dataset is designed to perform a stage-by-stage process of swapping four facial parts, namely nose, eyes, mouth and eyebrows, with the corresponding parts of the source image. The experimental setup requires all four parts to be edited, and all three face swapping methods are randomly applied to different parts to ensure diversity and complexity of the generated content. The final dataset contains 17440 image sets, each consisting of one original image and four images demonstrating the stage-by-stage editing process.
Based on the above challenges and observations, this paper proposes an innovative Modelship Attribution Transformer framework, which consists of three key components: the Sequence Slice, Self Spatial Extraction, and Modelship Trace Modeling. The framework is capable of generating Modelship Tokens by autoregressive means that accurately represent the models involved in the image during multiple editing processes and their execution sequences, thus achieving comprehensive attribution and traceability of complex generated content.

To summarize, the contributions of this work include:
\begin{itemize}
\item We introduce a new task, Modelship Attribution, which goes beyond traditional model attribution by not only identifying the generative model used to create synthetic content but also reconstructing the entire sequence of manipulations involved in its evolution. This task better aligns with the challenges presented by multi-stage content generation and manipulation in real-world scenarios.

\item To facilitate research on Modelship Attribution, we construct the Modelship Deepfake Dataset, comprising 17440 image sets generated through stage-by-stage editing using three fundamentally different face-swapping methods — StyleMapGAN \cite{kim2021exploiting}, DiffSwap \cite{zhao2023diffswap}, and FacePartsSwap \cite{ferrari2022makes}. This dataset reflects the complexity and diversity of real-world multi-stage manipulation scenarios, providing a benchmark for future studies.

\item We propose the Modelship Attribution Transformer (MAT) framework, which incorporates three novel components: Sequence Slice Module, Self Spatial Extraction Module and Modelship Trace Modeling Component. We conduct extensive experiments to demonstrate the effectiveness of our proposed MAT. The framework achieves high accuracy in reconstructing complex manipulation sequences, even under challenging settings involving diverse generative architectures and substantial content modifications.
\end{itemize}

\begin{figure*}[t]
  \centering
  \includegraphics[width=0.75\textwidth]{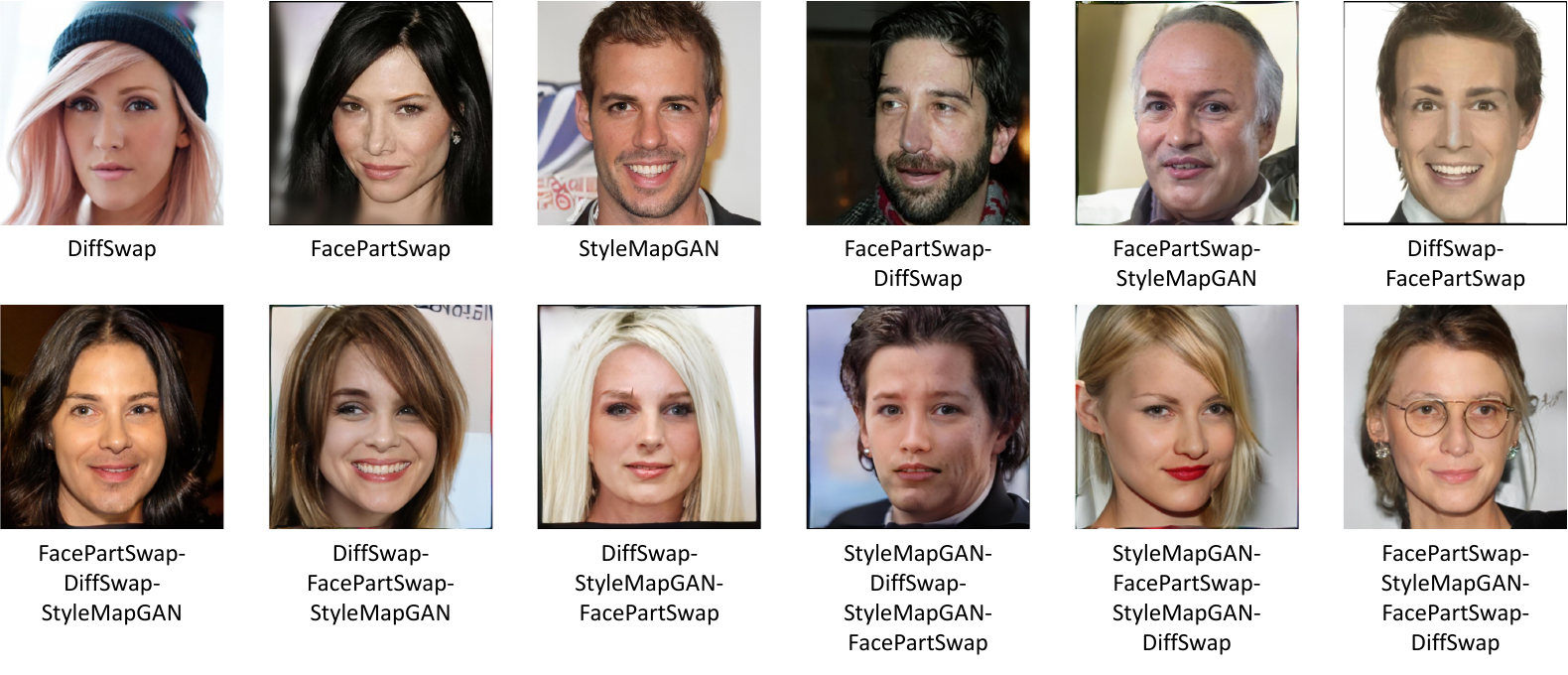}
  \caption{More examples from the modelship dataset. Each image sequence is labeled with the model used at each stage and the corresponding editing order, with facial regions edited randomly at each stage.}
  \Description{Examples}
  \label{fig:dataset-examples}
\end{figure*}
\section{Related Works}
\subsection{Model Attribution} 
Model attribution aims to identify the originating source model for given generative content and can be categorized into two branches, embedded fingerprinting and natural fingerprinting.  Embedded fingerprinting embeds “fingerprints,” also known as watermarks, into the generative model or its training data, ensuring they are distinct and easily identifiable \cite{nie2023attributing, yu2021artificial}.  For example, fingerprints can be deliberately designed and inserted during the model training phase, allowing for straightforward extraction from generated images to determine their source \cite{yu2020responsible}.  While effective, this method requires modifying the model architecture or training data, incurring significant overhead and limiting practicality in real-world applications. In comparison, natural fingerprinting does not require modifications to the models, datasets, or generation processes.  Instead, it relies solely on analyzing the final generated content, tracing "natural fingerprints" that unintentionally arise from the unique architectural or procedural characteristics of different models \cite{shahid2024generalized, sha2023fake}.  Current natural fingerprinting focuses on fine-grained attribution within the same lineage, such as GANs, but has limited generalizability to others, like diffusion models, which need extra effort.  For example, GANs carry distinct fingerprints even when trained with minimal differences in configuration, which allows for fine-grained and coarse-grained authentication \cite{yu2019attributing, yang2022deepfake}.  However, stable diffusion models pose unique challenges due to their latent space, which makes extracting fingerprints difficult \cite{wang2024did}.  Latent inversion-based methods, which are specifically designed for diffusion models, have the potential to simplify the attribution problem by focusing on the decoder and bypassing the complex latent denoising stage \cite{wang2024trace}.  Nevertheless, existing natural fingerprinting methods remain largely tailored to specific generative models and lack a unified framework for distinguishing between diverse architectures \cite{guarnera2024level}.

\subsection{Sequential Deepfake Detection}
Traditional deepfake detection is typically framed as a binary classification problem, where the task is to determine whether a given facial image is "real" or "fake". However, in real-world scenarios, facial images may undergo multiple sequential manipulations. For instance, a label such as "Eyebrow-Hair-Lip" indicates that the facial image has been manipulated in a specific order, modifying the eyebrows, hair, and lips sequentially. To address this complexity, sequential deepfake detection extends beyond binary classification by identifying not only the manipulated regions but also the precise order in which the manipulations occurred \cite{shao2022detecting}. This transforms the task into a ranking problem, making it significantly more challenging than standard detection methods \cite{hong2024contrastive}. 
To tackle this, Shao et al. proposed extracting the spatial relationships between facial components to identify manipulation traces and model their sequential relationships, enabling the detection of manipulation sequences \cite{shao2023robust}. Building on this, MMNet is the first network capable of recovering manipulated facial regions without requiring prior knowledge of the specific manipulation techniques \cite{xia2024mmnet}. Furthermore, Zhang et al. introduced a Spectral Transformer with a Pyramid Attention mechanism, which leverages spectral analysis and multi-scale attention to detect the permutation of manipulative operations in facial images \cite{zhang2024detecting}.

\afterpage{%
    \begin{figure*}[t]
        \centering
        \includegraphics[width=\textwidth]{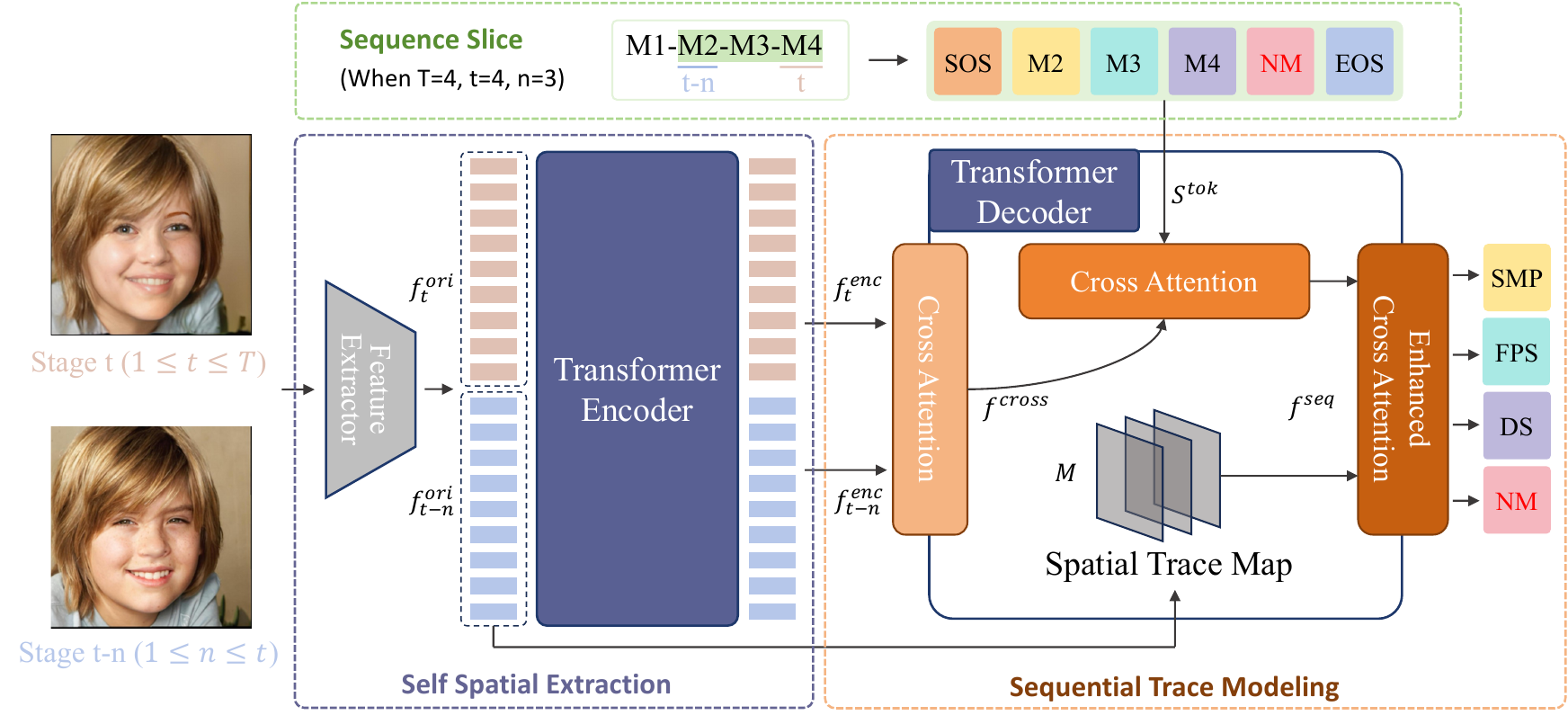}
        \caption{Overview of the proposed Modelship Attribution method. The approach tackles challenges in identifying editing paradigms by modeling generative processes across multi-stage manipulations. It consists of three components: Sequence Slice, which decomposes editing stages into transitions for analysis; Self Spatial Extraction, which enhances fine-grained features using CNN \cite{he2016deep}, FPN \cite{lin2017feature}, and a Transformer Encoder; and Modelship Trace Modeling, which employs Cross-Attention to detect model-specific traces and embeds operation sequences for feature interaction. A weight map highlights modified regions, enabling precise tracing and localization of editing stages.}
        \label{fig:Method}
    \end{figure*}
}
\section{Modelship Attribution}

\subsection{Benchmark Construction}

To investigate the traces left by different models when modifying specific regions of the same image, we selected three distinct face part-swapping methods based on different foundational frameworks: StyleMapGAN \cite{kim2021exploiting}, DiffSwap \cite{zhao2023diffswap}, and FacePartsSwap \cite{ferrari2022makes}. StyleMapGAN builds upon StyleGAN \cite{karras2019stylebasedgeneratorarchitecturegenerative} by introducing a spatially-aware latent space, known as StyleMap, and replacing adaptive instance normalization with spatially-varying modulation techniques. This enables more precise image generation and editing capabilities. DiffSwap, on the other hand, is a diffusion-based framework designed for high-fidelity and highly controllable face swapping. It reformulates the face-swapping task as a conditional inpainting problem guided by identity features and facial landmarks, achieving shape-consistent and identity-matched results in just two steps. FacePartsSwap utilizes 3D Morphable Models \cite{egger20203dmorphablefacemodels} combined with seamless cloning techniques to map source facial features onto target faces, allowing for detailed analysis of how different facial regions influence identity recognition.
All three methods enable precise control over specific facial regions, such as eyes, nose, or mouth, while preserving the overall context and integrity of the original image. Although DiffSwap is primarily designed for full-face swapping, it can leverage different masks during the inference process to selectively replace partial facial regions.

Since StyleMapGAN requires masks during the inference process, we selected the CelebAMask-HQ dataset \cite{lee2020maskgandiverseinteractivefacial}. The task of sequential facial components manipulation involves replacing specific facial regions from a source image with the corresponding regions in a target image. To study the natural fingerprints left by different generative models, the manipulation order of the four facial regions—nose, eyes, mouth, and eyebrows—was randomly shuffled for each image. Each of these four stages was assigned to a randomly selected deepfake model, and all three generative models (StyleMapGAN, DiffSwap, and FacePartsSwap) were required to participate in the process. ~\autoref{fig:dataset} illustrates two examples of images alongside their annotated operation workflows.
Since the combined processing of different models can sometimes lead to degraded results, with a small portion of the generated images appearing severely distorted, we applied the CLIB-FIQA \cite{CLIB-FIQA} method to evaluate the quality scores of the generated images. Based on these scores, we filtered out low-quality images to ensure that the dataset used for subsequent experiments maintains high reliability and consistency. Through this workflow, we ultimately generated 16,740 face part-swapping images across four stages of manipulation (including the original images), with annotations spanning 19 distinct labels. In total, the dataset contains 83,700 images (16,740 images × 5), providing a comprehensive resource for studying multi-stage manipulations and model attribution. More examples shown in \autoref{fig:dataset-examples}.

\subsection{Overview of Our Approach}
Based on the observations in Figure \ref{fig:Method}, we shift the task from extracting fingerprints of mixed models to identifying the editing paradigms of individual models. Specifically, the same model exhibits a consistent editing paradigm when applied to the same region, whereas different models or editing sequences lead to distinct results. However, accurately identifying these paradigms involves addressing several key challenges.
First, while different editing sequences can produce varying outputs, these differences are often subtle and confined to minute details, requiring heightened attention to fine-grained features in the images. Second, some models, such as StyleMapGAN, regenerate the entire image rather than only modifying the specified region. This regeneration can introduce subtle changes to non-targeted areas, potentially overwriting or distorting traces left by previous models in earlier stages, thereby increasing the complexity of detection. Lastly, models differ in the scope of their modifications: some models precisely edit only the specified regions using masks, while others may also introduce slight adjustments to areas like the background or hairstyle. These changes, whether prominent or subtle, must be carefully accounted for, with an appropriate weighting strategy applied to different regions to ensure the analysis remains robust and accurate.

To address the challenges discussed above, we propose a method called \textbf{Modelship Attribution}, which consists of three key components: \textbf{Sequence Slice}, \textbf{Self Spatial Extraction}, and \textbf{Modelship Trace Modeling}. This method aims to effectively model each editing stage and achieve precise tracing and localization of generative processes. 
To capture the influence of each editing stage on the image, we adopt an enhancement strategy utilizing intermediate results. Specifically, for a real image undergoing four editing stages by different models, we select the image from stage $t$ ($1 \leq t \leq 4$) and the image from stage $t-n$ ($1 \leq n \leq t$) as inputs. The goal is to learn the traces left by the model’s operation between stages $t-n$ and $t$. When $n=0$, the real image is used as the baseline. Consequently, the complete operation sequence is sliced into segments representing transitions from $t-n$ to $t$. These two input images are independently passed through a CNN and FPN \cite{lin2017feature} to extract rich detailed features, which are further refined using the Self-Attention mechanism in a Transformer Encoder to capture subtle traces left by the models during operations.
Next, the extracted features are processed by the Modelship Trace Modeling module. This module employs Cross-Attention to model the feature differences between the two input images, identifying the specific impact of the model’s operations. Additionally, the operation sequence is embedded into the feature interaction process to learn the correspondence between modification stages and feature changes. Finally, a weight map focusing on the modification regions is introduced to emphasize local feature variations, enabling efficient tracing and localization of each editing stage. This approach accurately captures operational paradigms and model traces, making it well-suited for complex multi-model editing tasks.

\subsection{Sequence Slice and Self Spatial Extraction}

In the Sequence Slice module, we decompose the sequential editing process into individual slice tasks, focusing on extracting the operational traces between specific editing stages. This slicing strategy effectively isolates the independent characteristics of each stage while avoiding feature entanglement.
Given an original image \(I_0\) that undergoes \(T\) sequential editing stages to produce the sequence \(\{I_1, I_2, \dots, I_T\}\), at each stage \(t\) (\(1 \leq t \leq T\)), we select the output image \(I_t\) at stage \(t\) and the image \(I_{t-n}\) from stage \(t-n\) (\(0 \leq n < t\)) as the input pair. The goal is to learn the operational traces of the model from stage \(t-n\) to stage \(t\). When \(n=0\), the real image \(I_0\) is used as the reference, and the slicing task captures the transition from the real image to the generated image at stage \(t\).
In addition to slicing the image sequence, the corresponding labels for the operations must also be sliced. Assuming the complete operational sequence is labeled as \(\text{M}_1, \text{M}_2, \dots, \text{M}_T\), corresponding to the models applied at each stage, the slicing operation extracts the relevant labels for the sliced image pair. For instance, if \(t=4\) and \(n=3\), the sliced input corresponds to stages \(\text{M}_2, \text{M}_3, \text{M}_4\). To indicate the end of operations and avoid ambiguity, any labels outside the sliced range are padded with \(\text{NM}\) (No Manipulation). Thus, the complete label sequence for this slice becomes \(\text{M}_2, \text{M}_3, \text{M}_4, \text{NM}\).

Since this task is essentially a sequence prediction problem, we further process the sliced operation sequence in an autoregressive manner to better capture the order of editing stages. Specifically, after the input sequence \(S = [M_{t-n}, \dots, M_t]\) is sliced, it is tokenized into fixed-dimensional representations. We add a Start of Sentence \(\text{[SOS]}\) token at the beginning and an End of Sentence \(\text{[EOS]}\) token at the end of the sequence to fully describe the sequence information. Hence, if the maximum number of editing stages is \(T=4\), the resulting tokenized sequence \(S^{\mathrm{tok}}\) has the shape \(d \times (T+2)\). Here, \(d\) represents the dimension of the tokenized feature embedding for each editing stage. Under this representation, the model can learn not only the individual features of each operation stage but also the dependencies between adjacent stages in the sequence.
The core of this slicing method lies in utilizing intermediate stages to decompose the entire sequence into localized feature transition tasks. This allows us to clearly capture the traces left by each model at specific stages while minimizing interference from multi-stage operations. The sliced images, \(I_t\) and \(I_{t-n}\), are passed through CNN \cite{he2016deep} and FPN \cite{lin2017feature} networks to extract fine-grained features, which are then utilized as inputs for subsequent Self Spatial Extraction and Modelship Trace Modeling modules.

Given input images \( I_t \) and \( I_{t-n} \) \(\in \mathbb{R}^{3 \times H \times W} \), where \( H \) and \( W \) represent the height and width of the image respectively, the image first passes through a feature extraction module. The CNN is used to capture local spatial patterns (e.g., textures and edges), while the FPN fuses multi-scale information from different resolution layers to produce feature maps that combine both fine details and global context. The extracted feature maps are denoted as \( f_t^{\text{ori}} \) and \( f_{t-n}^{\text{ori}} \in \mathbb{R}^{d \times h \times w} \), where \( d \) is the number of channels, and \( h \) and \( w \) correspond to the height and width of the feature maps, respectively.
To input the feature maps into a Transformer, they need to be reshaped into a sequence form \( \mathbb{R}^{d \times N} \), where \( N = h \times w \) is the sequence length. The Transformer encoder uses an attention mechanism to model long-range dependencies between different positions within these sequences.
In the encoder, the query vector \( Q \) and key vector \( K \) are generated by combining the input features \( f^{\text{ori}} \) with positional embeddings \( \text{pos} \). The value vector \( V \) directly uses the input features \({f}_{t}^{\text{enc}}, {f}_{t-n}^{\text{enc}} \) in perspective.
Finally, the outputs of multiple attention heads are concatenated and further processed to generate the final spatial features \({f}_{t}^{\text{enc}}, {f}_{t-n}^{\text{enc}} \in \mathbb{R}^{d \times N}\) from the encoder. These features contain rich contextual information, providing high-quality inputs for subsequent tasks.

\subsection{Modelship Trace Modeling}

Modelship Trace Modeling aims to model the feature differences between the input images and the temporal relationships in the operation sequence, thereby capturing subtle traces of model editing stages and enabling effective traceability. Specifically, we feed the spatial features \(f_t^{\text{enc}}\) and \(f_{t-n}^{\text{enc}}\) (extracted by the Transformer Encoder at stage \(t\) and \(t-n\)) into a Cross-Attention mechanism to capture their local feature differences. The purpose is to project the two sets of features into a shared space, allowing the model to focus on the specific change patterns triggered between stage \(t-n\) and stage \(t\).
We treat \(f_t^{\text{enc}}\) as the Query and \(f_{t-n}^{\text{enc}}\) as both the Key and the Value. The Cross-Attention mechanism outputs a set of cross-image feature representations \(f^{\mathrm{cross}}\), which characterize the local changes and feature differences when transitioning from \(f_{t-n}^{\text{enc}}\) to \(f_t^{\text{enc}}\).

Next, we feed both the tokenized operation sequence and \(f_{\text{cross}}\) into a Cross-Attention module to further strengthen the model’s understanding of the correspondence between the sequence of operations and the changes in image features through feature interaction. To more precisely capture the dynamic changes in local features, an additional Spatial Trace Map is introduced into the Cross-Attention module to guide the attention distribution. The Spatial Trace Map is a weight distribution map generated based on local changes in the input image, used to highlight the saliency of regions of interest and suppress irrelevant areas. Its purpose is to explicitly guide the attention to critical regions edited by the model, thus improving the model’s ability to capture the operation trajectory while reducing interference from background noise.

\begin{table*}[t]
\centering
\resizebox{0.47\textwidth}{!}{%
\begin{tabular}{l|cc|cc}
\Xhline{1pt}
\multirow{2}{*}{\textbf{Method}} & \multicolumn{2}{c|}{\textbf{ResNet-34}} & \multicolumn{2}{c}{\textbf{ResNet-50}} \\
\cline{2-5}
 & Ada-ACC & Strict-Match & Ada-ACC & Strict-Match \\
\hline
MC-Cls      & 60.94\% & -        & 61.45\% & -        \\
DRN \cite{wang2019detecting}        & -       & -        & 59.28\% & -        \\
DETR \cite{carion2020end}       & 62.37\% & -        & 61.96\% & -        \\
Two-Stream \cite{luo2021generalizing} & -       & -        & 64.73\% & -        \\
\hline
\textbf{Ours}        & \textbf{86.63\%} & \textbf{74.29\%}  & \textbf{87.57\%} & \textbf{76.01\%}  \\
\Xhline{1pt}
\end{tabular}%
}
\vspace{5pt}
\caption{Comparison of experimental results across different methods on ResNet-34 and ResNet-50 backbones, evaluated using Ada-ACC and Strict-Match metrics.}
\label{tab:results}
\end{table*}

\begin{table*}[t]
\centering
\begin{tabular}{c|cccc}
\Xhline{1pt}
\multirow{2}{*}{\textbf{Modules}}                                       & \multicolumn{2}{c|}{\textbf{ResNet-34}}               & \multicolumn{2}{c}{\textbf{ResNet-50}} \\ \cline{2-5} 
                                                               & Ada-ACC & \multicolumn{1}{c|}{Strict-Match} & Ada-ACC     & Strict-Match    \\ \hline
w/o Sequence Slice                                             & 75.98\%      & \multicolumn{1}{c|}{62.33\%}           & 77.61\%     & 64.48\%         \\ \cline{1-1}
w/o Sequence Slice, w/o Spatial Trace Map                      & 71.25\%      & \multicolumn{1}{c|}{57.98\%}           & 72.56\%     & 60.09\%         \\ \cline{1-1}
w/o Sequence Slice, w/o Spatial Trace Map, w/o Auto-regressive & 65.42\%      & \multicolumn{1}{c|}{-}            & 66.12\%     & -               \\ \Xhline{1pt}
\end{tabular}
\vspace{5pt}
\caption{Ablation study results for the Modelship Attribution Transformer (MAT). Removing the Sequence Slice or Spatial Trace Map reduces accuracy, highlighting their importance for capturing fine-grained features and spatial information. Replacing the auto-regressive approach with multi-label classification further lowers performance, showing the critical role of each component.}
\label{tab:Ablation}
\end{table*}

We first utilize the feature maps \(f_t^{\mathrm{ori}} \) and \(f_{t-n}^{\mathrm{ori}} \in \mathbb{R}^{d \times h \times w}\) extracted by the CNN+FPN in the Self Spatial Extraction module. We then compute the absolute difference between these two feature maps along the channel dimension to obtain the difference tensor \(\Delta f = |f_t^{\mathrm{ori}} - f_{t-n}^{\mathrm{ori}}|\). 
To simplify subsequent stages and highlight the most significant variation at each pixel, we take the maximum value along the channel dimension, obtaining an initial 2D difference map \(\Delta f_{\mathrm{map}} = \max(\Delta f, \text{axis} = d)\). However, directly using this 2D map may lead the model to focus excessively on large-magnitude changes while overlooking smaller yet potentially meaningful variations. Therefore, we normalize \(\Delta f_{\mathrm{map}}\) so that all pixel values lie in the range \([0, 1]\):  
\[
\Delta f_{\mathrm{norm}}(h, w) = 
\frac{\Delta f_{\mathrm{map}}(h, w)}{\max(\Delta f_{\mathrm{map}})}.
\]
After obtaining the normalized difference map \(\Delta f_{\mathrm{norm}}\), we use an MLP to predict the center \((t_h, t_w)\) and the scale parameters \((r_h, r_w)\) of the changing region. Based on these parameters, we generate a Gaussian-shaped Spatial Trace Map:
\[
M(h, w) = 
\Delta f_{\mathrm{norm}}(h, w) \cdot 
\exp\Bigl(
  -\frac{(h - t_h)^2}{\lambda r_h^2}
  -\frac{(w - t_w)^2}{\lambda r_w^2}
\Bigr),
\]
where \(\lambda\) is a hyperparameter controlling the spread of the Gaussian distribution. By applying a Gaussian weighting mechanism, the Spatial Trace Map not only emphasizes the most prominent areas of change between the images but also allocates additional focus to subtler variations that might be crucial.
With this Spatial Trace Map \(M\) computed, we embed it into the attention computation as an explicit prior. Let \(Q, K, V\) be the projected features from the tokenized operation sequence \(S^{\mathrm{tok}}\) (for \(Q\)) and the cross-image features \(f_{\text{cross}}\) (for \(K\) and \(V\)). We can write the single-form Cross-Attention as:
\[
f^{\text{seq}}
=
\mathrm{Softmax}\Bigl(
  \frac{(Q\,W_Q)\,(K\,W_K)^\top}{\sqrt{d_k}}
  + \log M
\Bigr)
\,(V\,W_V).
\]

where \(W_Q, W_K, W_V\) are linear projection matrices, and \(\log M\) explicitly highlights important regions while suppressing irrelevant areas.
By integrating the Spatial Trace Map into the attention computation, the model can more accurately align the tokenized operation sequence with the changes in image features. This mechanism ensures that crucial regions—those that have undergone edits—are explicitly emphasized, which in turn enhances the model’s ability to capture the trajectory of editing operations while mitigating background noise interference.

\section{Experiments}
\subsection{Experimental Setup}
{\bfseries Datasets.} 
We conducted experiments on the Modelship dataset, a custom-created dataset containing 16,740 images representing four stages of a process. The dataset was split into training (80\%), validation (10\%), and testing (10\%) subsets. It features three label types—StyleMapGAN, DiffSwap, and FacePartsSwap—forming a total of 19 permutations. Random edits were applied to four facial regions: nose, eyes, mouth, and eyebrows, with all images resized to 256×256 pixels.

{\bfseries Implementation Details.} 
For the models, we used ResNet-34 and ResNet-50 CNNs \cite{he2016deep} pre-trained on the ImageNet dataset \cite{deng2009imagenet}, alongside a transformer model designed with 2 encoder layers, 2 decoder layers, and 4 attention heads. To ensure comparable parameter sizes, the transformer and CNNs were trained under consistent conditions. The training schedule involved 150 epochs, with an initial learning rate of \(1e^{-3}\) for the transformer and \(1e^{-4}\) for the CNNs. The learning rate was reduced by a factor of 10 at epochs 70 and 120. We used a batch size of 64 and set the weight parameter \(\lambda\) = 4 to balance the training objectives. All experiments are implemented with Pytorch on 4 NVIDIA Geforce RTX 3090s.

\subsection{Baseline Comparison}
Although our method adopts an autoregressive approach to predict manipulation sequences, a more straightforward strategy is to frame this task as a multi-class classification problem with 19 categories.  As a baseline, we design a simple multi-class classification network MC-Cls that learns to predict multiple manipulations. Methods such as DRN \cite{wang2019detecting}, DETR \cite{carion2020end}, and Two-Stream \cite{luo2021generalizing} treat each manipulation sequence as a separate class, transforming deepfake detection into a multi-class classification task. These methods, like ours, utilize ResNet-34 and ResNet-50 pretrained on the ImageNet dataset as their backbone networks. However, directly defining the task as multi-class classification may encounter challenges such as class imbalance and insufficient modeling of the complex dependencies between sequences. 
\begin{figure}[t]
  \centering
  \includegraphics[width=\linewidth]{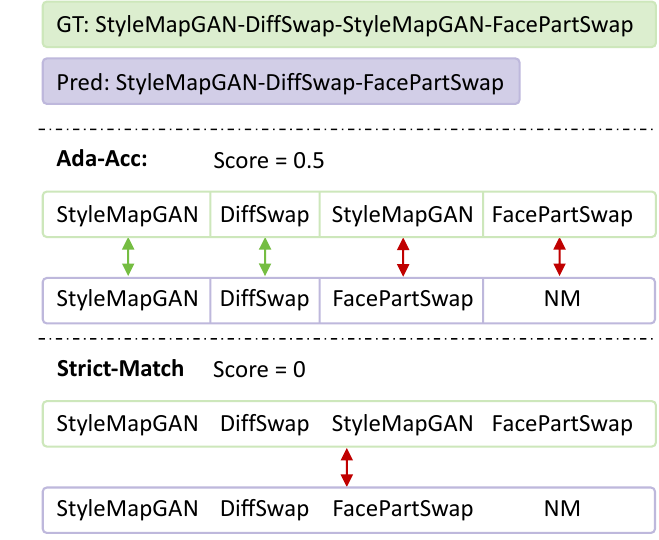}
  \caption{Illustration of evaluation metrics. Ada-ACC aligns sequences with padding for one-to-one token comparison, calculating accuracy as the ratio of correct matches. Strict-Match requires exact sequence matching, with any mismatch yielding 0\% accuracy.}
  \Description{Examples}
  \label{fig:Metrics}
\end{figure}

{\bfseries Evaluation Metrics.}
The evaluation metrics are illustrated in Figure~\ref{fig:Metrics}. In the adaptive accuracy (Ada-ACC) evaluation metrics, the process begins by aligning the predicted sequence and the ground truth sequence at their respective start points using a special [SOS] token, ensuring the sequences are compared from a common reference point. Following this, the sequences are adjusted to have equal lengths by appending a "No Manipulation" (NM) token to the shorter sequence immediately before its [EOS] token. This padding process ensures that both sequences are of identical length and enables a direct one-to-one comparison of their respective elements.
Once the sequences are aligned and matched in length, the evaluation proceeds by comparing tokens at each position in the two sequences. Tokens are considered correct if they match exactly in both sequences at the same position, with the order of tokens strictly respected. Any mismatch between tokens is classified as an error. Finally, the accuracy score is calculated as the ratio of correctly matched tokens to the total number of tokens in the padded sequence.  

For comparison with multi-class classification tasks, we can introduce another “Strict-Match” evaluation method whereby, once the sequences have been aligned and padded to equal length, any single mismatch between the predicted sequence and the ground truth immediately deems the prediction incorrect; in other words, the sequence is considered correct only if every token exactly matches its counterpart, yielding an accuracy of 100\%, and otherwise 0\%.



{\bfseries Results.} 
\autoref{tab:results} presents the performance of different methods on ResNet-34 and ResNet-50 architectures, evaluated using the Ada-ACC and Strict-Match metrics. Since ML-Cls, DRN, DETR, and Two-Stream are designed as multi-label classification tasks, they do not have Strict-Match accuracy scores. Due to their lack of capability in capturing detailed modelship information, their detection performance is significantly lower than that of our proposed method. Moreover, as more models are introduced and the sequence of operations becomes more complex, these classification methods require an increasing number of classification branches, leading to a surge in model parameters and a decline in performance.

Our proposed Modelship Attribution Transformer (MAT) achieves significantly better results in terms of the Ada-ACC metric, surpassing other methods by up to 28.29\% on the ResNet-50 backbone. To ensure fair comparisons with multi-label classification tasks, we also introduced the Strict-Match evaluation metric. Notably, our method achieves an impressive Strict-Match accuracy of 76.01\%, which exceeds other methods by up to 16.73\%.
These results clearly demonstrate the superior performance of MAT. The proposed method effectively captures the editing paradigms of different models, addressing the challenge of overwritten model fingerprints and showcasing robust and precise performance even in more complex tasks.

\subsection{Ablation Study}
In the ablation study, removing the Sequence Slice—i.e., directly feeding only the final M4 output to the transformer for Enhanced cross attention with \(S^{tok}\)—results in a notable drop in performance as \autoref{tab:Ablation} shows, suggesting that leveraging intermediate features is crucial for capturing fine-grained visual cues. Eliminating the Spatial Trace Map from the final Enhanced cross attention causes an additional decline, highlighting the value of explicit spatial information. Finally, discarding the auto-regressive approach converts the problem into a multi-label classification task, which not only further reduces accuracy but also renders the Strict-Match metric inapplicable. Collectively, these experiments illustrate that each component plays a significant role in boosting the model’s prediction accuracy and robustness.

\section{Conclusion}
In conclusion, our proposed task of Modelship Attribution addresses the pressing challenge of tracking multi-stage manipulations from diverse generative models. By moving beyond traditional single-model identification, our approach reconstructs the complete editing sequence and the specific models involved at every stage, enabling thorough traceability of manipulated content. To facilitate research on this complex problem, we introduce the Modelship deepfake dataset, which includes multiple editing stages by three distinct face-swapping methods (GAN-based, diffusion-based, and 3D reconstruction-based), reflecting real-world scenarios. Finally, we present the Modelship Attribution Transformer framework, featuring the Sequence Slice, Self Spatial Extraction, and Modelship Trace Modeling modules that effectively capture the intricate evolution of manipulations. Our experiments demonstrate its capability to generate Modelship Tokens in an autoregressive manner, paving the way for more reliable and comprehensive detection, attribution, and security measures in the rapidly evolving domain of AI-generated content.

\bibliographystyle{ACM-Reference-Format}
\bibliography{sample-base}

\end{document}